\documentclass[sigconf]{acmart}

\usepackage{array}
\usepackage{amssymb}
\usepackage{fontawesome}
\usepackage{multirow}
\usepackage[utf8]{inputenc}

\AtBeginDocument{%
  \providecommand\BibTeX{{%
    \normalfont B\kern-0.5em{\scshape i\kern-0.25em b}\kern-0.8em\TeX}}}

\copyrightyear{2020}
\acmYear{2020}
\setcopyright{acmlicensed}
\acmConference[IUI '20]{25th International Conference on Intelligent User Interfaces}{March 17--20, 2020}{Cagliari, Italy}
\acmBooktitle{25th International Conference on Intelligent User Interfaces (IUI '20), March 17--20, 2020, Cagliari, Italy}
\acmPrice{15.00}
\acmDOI{10.1145/3377325.3377538}
\acmISBN{978-1-4503-7118-6/20/03}

\begin{document}

\title{AutoAIViz: Opening the Blackbox of Automated Artificial Intelligence  with Conditional Parallel Coordinates}

\author{
    Daniel Karl I. Weidele*, 
    Justin D. Weisz,
    Erick Oduor,
    Michael Muller,
}
\author{
    Josh Andres,
    Alexander Gray and 
    Dakuo Wang
}
\affiliation{%
 \institution{IBM Research~-~daniel.karl@ibm.com}
}

\renewcommand{\shortauthors}{Daniel Karl I. Weidele et al.}

\renewcommand{\authors}{Daniel Karl I. Weidele, Justin D. Weisz, Eno Oduor, Michael Muller, Josh Andres, Alexander Gray and Dakuo Wang}

\newcolumntype{P}[1]{>{\raggedright\arraybackslash}p{#1}}

\newtheorem{axiom}{Axiom}

\newtheorem{hypothesis}{H}



\begin{abstract}

Artificial Intelligence (AI) can now automate the algorithm selection, feature engineering, and hyperparameter tuning steps in a machine learning workflow. Commonly known as AutoML or AutoAI, these technologies aim to relieve data scientists from the tedious manual work. However, today's AutoAI systems often present only limited to no information about the process of how they select and generate model results. Thus, users often do not understand the process, neither do they trust the outputs. In this short paper, we provide a first user evaluation by 10 data scientists of an experimental system, AutoAIViz, that aims to visualize AutoAI's model generation process. We find that the proposed system helps 
users to complete the data science tasks, and increases their understanding, 
toward the goal of increasing trust in the AutoAI system.

\end{abstract}



\maketitle

\section{Introduction}
Data scientists apply machine learning techniques to gain insights from data in support of making decisions~\cite{kross2019practitioners}. The challenge is that this process is labor intensive, requiring input from multiple specialists with different skill sets~\cite{yang2018grounding, muller2019datascience, amershi2019software,Mao_2019, amyzhang2020}. As a result, AI and Human-Computer Interaction (HCI) researchers have investigated how to design systems with features that support data scientists in creating machine learning models~\cite{lee2019human, kery2018story, rule2018exploration, wang2019humanai, wang2019data, muller2019datascience}. This work is often referred to as \textbf{human-centered machine learning} or \textbf{human-in-the-loop machine learning}~\cite{lee2019human, gil2019towards, wang2019humanai}.

Recently, techniques have been developed to automate various steps of the data science process, such as preparing data, selecting models, engineering features, and optimizing hyperparameters~\cite{liu2019admm,khurana2016cognito, kanter2015deep,lam2017one, zoller2019survey}. These techniques -- known as \textbf{AutoAI} or \textbf{AutoML} -- are being incorporated into commercial products (e.g.~\cite{web:googleautoml, web:microsoftazure,web:ibmautoai, web:h2o, web:datarobot}) as well as open source packages~\cite{web:tpot, olson2016tpot, web:autosklearn}. Although the ability for AutoAI to speed up the model-building process is very promising, there are challenges due to the opaque nature of how AutoAI actually creates models~\cite{wang2019atmseer, wang2019humanai, lee2019human}. Users do not necessarily understand \emph{how} and \emph{why} AutoAI systems make the choices they make when selecting algorithms or tuning hyperparameter values. This ``blackbox'' nature of AutoAI operation hinders the collaboration between data scientists and AutoAI systems~\cite{wang2019atmseer}.



Several solutions have been proposed to make AutoAI more transparent in its operation in order to increase trust. These proposals include documenting the process by which machine learning models were created~\cite{hind2018increasing, mitchell2019model} and using visualizations to show the search space of candidate algorithms or hyperparameter values~\cite{wang2019atmseer,golovin2017google,park2019visualhypertuner}. However, these solutions either document the state of an AI model \emph{after} it has been created, or focus only on \emph{one step} of the model generation workflow (e.g., search algorithms or selecting hyperparameters). There remains a gap in providing users with an overview of how an AutoAI system works \emph{in operation}, from the moment data are read to the moment candidate models are produced.


In this paper, we provide a first user evaluation of a
new kind of visualization --
\textbf{Conditional Parallel Coordinates (CPC)}~\cite{DBLP:journals/corr/abs-1906-07716} -- to open up the ``black box'' of AutoAI operation. CPC is a variation of the parallel coordinates visualization~\cite{inselberg1990parallel, inselberg2009parallel}, 
that allows
the user to expose increasing amounts of detail added to a conventional Parallel Coordinates display
interactively. 
We integrated CPC into an existing AutoAI platform~\cite{wang2019humanai} and conducted a usability study with 10 professional data scientists to understand its usefulness. 
Our results suggest that data scientists are able to successfully glean information from the visualization in order to gain a better understanding of how AutoAI makes decisions, while those models are being built. Our work shows how increased \emph{process} transparency leads to improved collaboration between data scientists and AutoAI tools, 
as a 
step 
toward democratizing AI for non-technical users.
\section{Related Work}

\begin{figure*}[ht]
    \centering
    \includegraphics[width=0.8\linewidth]{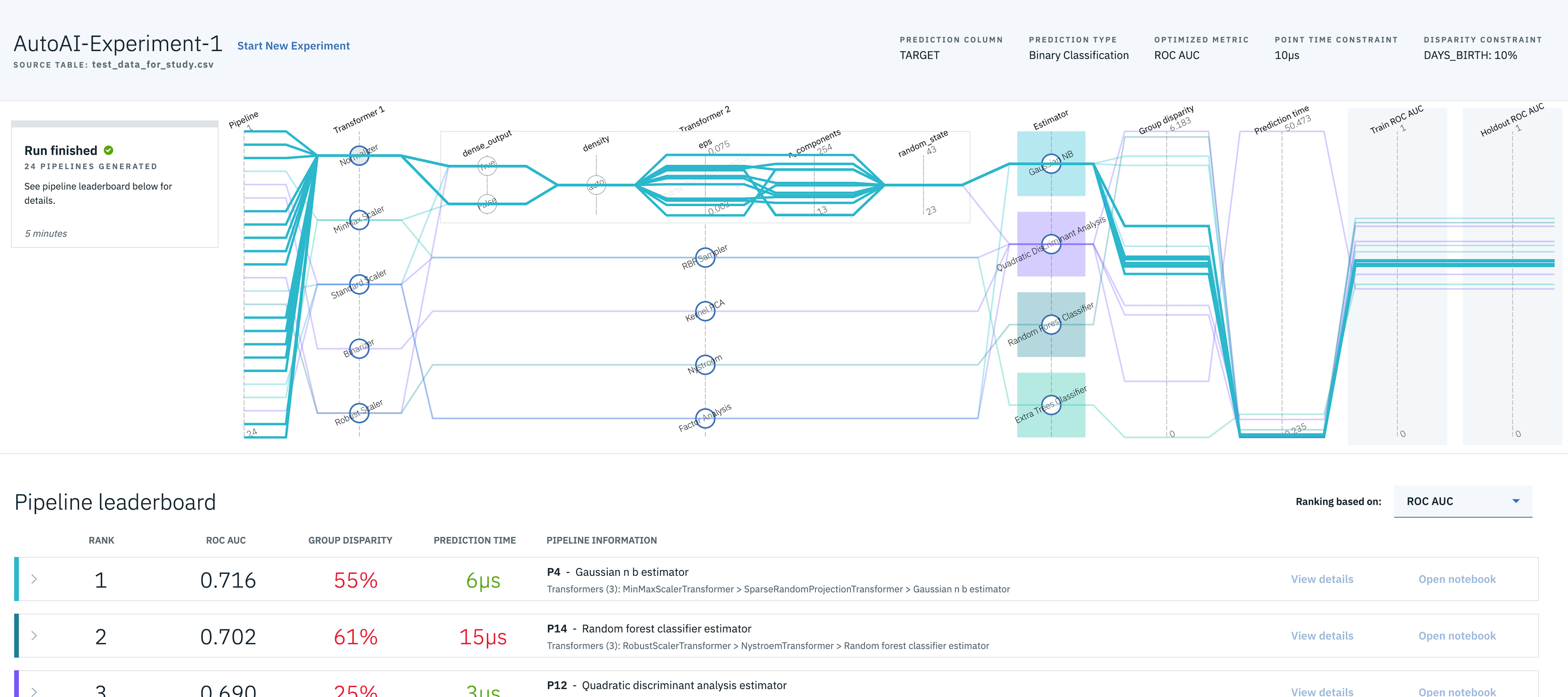}
    \caption{Screenshot showing the CPC visualization (top) and the leaderboard (bottom, partially shown). In CPC, each colored line running from left to right represents a machine learning pipeline, and corresponds to a row in the leaderboard. In this example, all pipelines consist of three steps: transformation step 1, transformation step 2, and estimator. Pipelines are evaluated on four metrics: group disparity, prediction time, and ROC AUC scores on the training data and the holdout data.}
    \label{fig:cpc}
\end{figure*}



\subsection{Human-in-the-loop Data Science and Automated Machine Learning}
Many studies have focused on understanding the work practices of data scientists. Muller et al. and Hou \& Wang both reported that data science workflows operate as an iterative conversation between data scientists and other stakeholders, focused around data and model artifacts~\cite{muller2019datascience,hou2017hacking}. As such, data science work is a confluence of people with many different roles, skills, and needs~\cite{amyzhang2020}. 
Amershi et al. 
highlighted the importance of adopting a human-centered approach to building data science tools~\cite{amershi2019guidelines, amershi2011human}. 
Kross \& Guo
users expressed
a strong desire 
for an integrated user interface with both code and narrative.
These needs are perhaps best captured in the narrative uses \cite{kery2018story} of the Jupyter Notebook environment~\cite{web:jupyter, web:jupyterlab}, and researchers have conducted numerous studies of how data scientists incorporate notebooks into their workflows~\cite{rule2018exploration,passi2018trust}, how they conduct version control for notebooks~\cite{kery2019towards}, and how they enable simultaneous multi-user editing in notebooks~\cite{wang2019data}.



What remains less well-understood is how data scientists will incorporate automated machine learning technologies into their workflows. While initial work suggests that data scientists see AutoAI tools as collaborative~\cite{wang2019humanai}, much about the underlying process by which AutoAI operates remains obscure to data scientists~\cite{wang2019atmseer, golovin2017google, park2019visualhypertuner, lee2019human}. Our efforts in this work focus on increasing trust in AutoAI by increasing the transparency of its operation.

\subsection{Trust via Transparency in AI Systems}

Recent criticisms of ``blackbox'' machine learning models (e.g.~\cite{rudin2018please,liu2019admm,liao2020questioning,zhang2020effect}) begin with the notion that if one does not understand \emph{how} a model produces a recommendation, then that recommendation may not be trustworthy. The issue of trust may be crucial for AutoAI systems~\cite{JaimieDrozdal2020}. Not only must a model's recommendations be explainable, but also the process by which the model was created. 




Visualizations provide a promising way for increasing transparency of operation, and several systems have already been created that focus on specific points within the data science workflow. Vizier~\cite{golovin2017google} and VisualHyperTuner~\cite{park2019visualhypertuner} both focus on hyperparameter tuning, showing the relationships between a model's hyperparameters and performance. 
ATMSeer~\cite{wang2019atmseer} provides a multi-granularity visualization of model selection and hyperparameter tuning, enabling users to monitor the process and adjust the search space in real time. 
While these visualizations improve transparency of single components in the AutoAI process, no visualization has yet tackled a representation of the entire process, from data ingestion to model evaluation. To visualize the entire process, we use Conditional Parallel Coordinates (CPC), which offers data scientists, (a) a real-time overview of the machine learning pipelines as they are being created, and (b) the ability to view detailed information at each step in the AutoAI process.

\section{Conditional Parallel Coordinates for AutoAI}


We provide a first user\--evaluation of a recently proposed Conditional Parallel Coordinates (CPC) component \cite{DBLP:journals/corr/abs-1906-07716}. Compared to classic Parallel Coordinates (PC) visualizations~\cite{gannett1883general, inselberg1985plane}, CPC introduces new layers of details recursively.


Pipeline configuration data generated by an AutoAI search process is considered conditional data. We say hyperparameters are conditioned on the choice of a pipeline step. This conditionality enables CPC, which allows to inspect conditional layers. AutoAI first chooses a sequence of transformation steps before training an estimator. Then, the hyperparameters of these choices are fine-tuned. Further, the user can bound the AutoAI search space by setting constraints, such as \emph{group disparity} against model bias, or \emph{prediction runtime} for inference speed~\cite{liu2019admm}. Typically, during the process pipelines are being optimized towards a performance measure, in our case ROC AUC. 

Figure~\ref{fig:cpc} shows our suggested mapping of this data to CPC. On the top level we generate an axis for each of the following attributes: \textit{Pipeline ID, Transformer 1, Transformer 2, Estimator, Group Disparity, Prediction Runtime}, and \textit{ROC AUC} on training and holdout split respectively. In the next level of detail, for categorical choices, we construct a conditional axis per hyperparameter. The figure shows the visualization after the user expanded the Sparse Random Projection Transformer in the Transformer 2 step. Here, the visualization further allows to differentiate pipelines across the 5 hyperparameters \emph{dense\_output, density, eps, n\_components} and \emph{random\_state}.

Thus, in our CPC visualization users have the ability to drill down into a particular step of the machine learning pipeline by clicking it. This reveals further details about this step, which in turn are again visualized in the form of PC. We assume this recursive process together with seamless preservation of intuitive interaction patterns from PC will enable users to familiarize themselves with the limited additional complexity rather quickly.

Next to the CPC component, towards the bottom of the interface we feature a pipeline leaderboard, showing pipeline configuration data sorted according to ROC AUC descendingly. For each pipeline row, from left to right, we display the rank, ROC AUC, Group Disparity, Prediction Time, ID and information on the steps.

Color indicates the choice of the estimator, as well for pipelines in CPC as for rows in the leaderboard. In principle the color mapping could be left up to the user to support focusing on different axes, however, for this study we intended to limit further complexities.

Lastly, AutoAIViz is capable of reflecting the AutoAI search status live, i.e. as the search progresses in the background latest pipelines are added to the visualizations.

\section{User-Evaluation Study}



To evaluate the CPC visualization in the context of AutoAI, we recruited 10 professional data scientists across three continents. We developed a set of 17 informational questions for participants to answer using AutoAIViz (Table \ref{tab:study}), based on a binary classification task. Each question asked about one factual piece of information contained in the interface. Although all questions could be answered with the CPC visualization alone, we included the Leaderboard visualization (shown at the bottom of Figure~\ref{fig:cpc}) in order to represent existing AutoAI user interfaces, and gauge which representation was preferable for participants. Our questions included finding the total number of pipelines generated by AutoAI (Q1), finding the training data ROC AUC score for a given pipeline (Q9), and finding the most frequently searched transformer (Q13).


Participation in our evaluation took approximately 30 minutes.  Participants were allowed a few minutes to freely explore the interface before they were presented with the set of questions. Due to our desire to gauge CPC's self-explanatory nature, participants were given minimal explanation about the visualization or how to interact with it. Participants worked for about 10-15 minutes in answering the questions. As in other studies of visualizations (e.g.~\cite{chittaro2004visual, holliday2016user,Wang:2015:DVC:2702123.2702517,dakuo}), participants were encouraged to think aloud as they explored the interface and ask any questions they had.

We tracked how participants answered each question, either by consulting the CPC visualization (C), the Leaderboard (L), or both (B). All but one of our questions were factual, and thus had either correct and incorrect answers; one question asked participants to select a single pipeline and then answer 4 questions specific to it.

In addition to the questions, we had participants self-assess their \emph{familiarity} with, \emph{understanding} of, and \emph{trust} in AutoAI systems on 5-point Likert scales. Familiarity was assessed prior to the study; understanding and trust were measured at the beginning and the end. Finally, we asked participants to reflect on elements they liked and disliked during the study, as well as their desire to use AutoAI in the future as part of their work practice.

We primarily examine the extent to which participants were able to correctly find each piece of information and which visualization they used to find it, as well as their understanding of and trust in AutoAI changed in the study.

\section{Results}


\subsection{Ability to Interpret CPC}

Across all 17 questions, we find that CPC has been used three times more often than the leaderboard
($\chi^2_1 = 42.050$, p<.001). However, to learn about the results on the components more thoroughly, let us compare which of the two components has been favorably consulted depending on the types of questions.

Questions on total counts, but also counts \emph{grouped by} transformers or estimators have mostly been answered using CPC. Further, all candidates were able to intuitively learn how to reveal hyperparameter information, with only 2 people trying to look for the answer to such questions in the leaderboard. Interestingly, even question 17, concerned with the difference between the best and the worst pipeline, has been answered favorably using CPC only, making participants feel comfortable enough to avoid a scroll to the far end of the leaderboard. One participant also managed to answer all questions correctly solely based on CPC.

We further find the leaderboard serves a differentiating purpose. For questions where a single pipeline needed to be identified based on the result metrics, participants preferred to consult the leaderboard. This is particularly the case for Q6, where we put our participants into a conflicting scenario by asking for the best pipeline \emph{in their opinion}. 40\% would not overwrite the top pipeline of the leaderboard. 50\% opted for a slightly less accurate pipeline roughly cutting group disparity in half and predicting twice as fast. Only one subject opted for a pipeline with group disparity of 0\%, setting back ROC AUC to 0.500. For this question not a single participant based their decision on CPC only. Surprisingly, over the course of the next 3 questions where only single data of the \emph{just chosen} pipeline row was asked for, up to 6 participants hovered back to instead read the results from the CPC visualization. Overall, we further leave with the impression that subjects did not have difficulties switching between the components.

With respect to the quality of provided answers, across all questions, we were not able to determine a statistically significant dependency between the question outcome (correct / incorrect) and the used component ($\chi^2$ = 0.42, p < .81), which means CPC did not introduce systematic error compared to the leaderboard.

Regarding the questions most frequently answered incorrectly, some participants found the mix of pipeline steps with result metrics confusing in the beginning (Q2). We believe this can be improved visually by separating the axes more clearly, or adding tool tips. Q11 is asking for the top estimator, however, participants slipped and reported the top pipeline. This seems more of an issue with the study than the components themselves. Up to 40\% struggled giving correct answers on hyperparameter questions (Q15, Q16), ignoring numerical hyperparameter axis, or counting values instead of axis.







\subsection{Effect of CPC on Trust and Understanding}
Participants reported some initial familiarity with AutoAI prior to the study (M (SD) = 2.5 (1.51) of 5). 
Ratings of trust fell in the middle of the scale prior to using CPC, with a non-significant increase after using CPC (pre: M (SD) = 2.7 (1.06) of 5, post: M (SD) = 3.3 (1.06), t(9) = 1.77, p < .11). By contrast, understandability increased significantly (pre: M (SD) = 2.4 (0.97), post: M (SD) = 3.6 (0.52), t(9) = 3.67, p < .005). We speculate that, with a larger sample size, the difference in trust might also have achieved significance.




\subsection{Qualitative Feedback}
After the study participants were further asked to give their subjective opinion on what they liked or disliked in AutoAIVIz. 

On the positive side, 8 out of the 10 participants found the interactive CPC visualization useful, mostly for its ability to simultaneously display pipeline steps, hyperparameters and results. The remaining 2 candidates more commended the overall system, specifically as a way to generate competitive baselines with little effort.

On the negative side, candidates sometimes found the CPC display to be crowded, thus making it harder to perceive individual or subsets of pipelines, or hyperparameters. Participants were also not always familiar with the terminologies used in AutoAIViz. Lastly, 2 subjects wished they were given more power to fine-tune the AutoAI manually, i.e. through code.

We also wanted to know if participants could imagine AutoAIViz as part of their job in the future. Half of our participants could clearly see themselves work with AutoAIViz after some minor improvements, with one candidate even already relying on a similar system. With only a single subject arguing against AutoAI correlated to lacking trust, the remaining four expressed they would consult AutoAI moderately, as a baseline to kick-off new projects, or in cases when they are less informed.

{\small
\begin{table}
\caption{Quantitative results of the study. In case of multiple answers we highlight the correct answer bold. Note {\small \textbf{[Q6]}} is subjective, so all answers are correct. For {\small \textbf{[Q6-10]}} we provide aggregated results of correct (\checkmark) and incorrect answers ($\times$).}
\label{tab:study}
\resizebox{\columnwidth}{!}{%
\begin{tabular}{P{7cm}crrrrr}
    \toprule
    & Result & \# & \#C & \#L & \#B \\
    \midrule
    
    \textbf{Q1} Find the total number of pipelines generated by AutoAI & 24 & 10 & 6 & 0 & 4  \\
    
    \midrule
    \multirow{2}{*}{\parbox{7cm}{\textbf{Q2} Find the total number of steps in the pipelines}} & \textbf{3} & 6 & 3 & 1 & 2  \\
    \cline{2-6}
    & 7 & 4 & 4 & 0 & 0 \\
    
    \midrule
    \textbf{Q3} Find the tot. num. of transf. evaluated in transf. step 1 & 5 & 10 & 10 & 0 & 0  \\
    
    \midrule
    \textbf{Q4} Find the tot. num. of transf. evaluated in transf. step 2 & 5 & 10 & 10 & 0 & 0  \\
    
    \midrule
    \textbf{Q5} Find the tot. num. of estim. evaluated in the estim. step & 4 & 10 & 10 & 0 & 0  \\
    
    \midrule
    \multirow{3}{*}{\parbox{7cm}{\textbf{Q6} Find the best pipeline in your opinion}} & \textbf{P12} & 5 & 0 & 4 & 1  \\
    \cline{2-6}
    & \textbf{P4} & 4 & 0 & 1 & 3 \\
    \cline{2-6}
    & \textbf{P21} & 1 & 0 & 1 & 0 \\
    
    \midrule
    \textbf{Q7} Find the group disparity score of this pipeline (Q6) & \checkmark & 10 & 2 & 7 & 1 \\
    
    \midrule
    \textbf{Q8} Find the prediction time of this pipeline (Q6) & \checkmark & 10 & 3 & 6 & 1 \\
    
    \midrule
    \textbf{Q9} Find the Train. data ROC AUC score of this pipeline (Q6) & \checkmark & 10 & 6 & 3 & 1 \\
    
    \midrule
    \multirow{2}{*}{\parbox{7cm}{\textbf{Q10} Find the Holdout data ROC AUC score of this pipeline (Q6)}} & \checkmark & 9 & 5 & 0 & 4   \\
    \cline{2-6}
    & $\times$ & 1 & 0 & 1 & 0 \\
    
    \midrule
    \multirow{2}{*}{\parbox{7cm}{\textbf{Q11} Find the best performing estimator regarding Training ROC AUC performance score}} & \textbf{GNB} & 6 & 3 & 1 & 2  \\
    \cline{2-6}
    & P4 & 3 & 2 & 1 & 0 \\
    \cline{2-6}
    & QDA & 1 & 0 & 1 & 0 \\
    
    \midrule
    \multirow{2}{*}{\parbox{7cm}{\textbf{Q12} Is there a pipeline that satisfies both constraints?}} & \textbf{No} & 9 & 1 & 4 & 4  \\
    \cline{2-6}
    & Yes & 1 & 1 & 0 & 0 \\
    
    \midrule
    \multirow{2}{*}{\parbox{7cm}{\textbf{Q13} Find the most frequently searched transformer}} & \textbf{SRP} & 8 & 8 & 0 & 0  \\
    \cline{2-6}
    & Norm. & 2 & 2 & 0 & 0 \\
    
    \midrule
    \textbf{Q14} Find the most frequently searched estimator & GNB & 10 & 9 & 0 & 1  \\
    
    \midrule
    \multirow{3}{*}{\parbox{7cm}{\textbf{Q15} Find the total number of hyperparameters of Quadr. Discriminant Analysis Estimator}} & \textbf{5} & 6 & 6 & 0 & 0  \\
    \cline{2-6}
    & 4 & 2 & 1 & 0 & 1 \\
    \cline{2-6}
    & 2 & 2 & 2 & 0 & 0 \\
    
    \midrule
    \multirow{3}{*}{\parbox{7cm}{\textbf{Q16} Find the hyperparameter that most influenced the Quadr. Discr. Analysis Estimator}} & \textbf{tol} & 7 & 6 & 0 & 1  \\
    \cline{2-6}
    & N/A & 2 & 1 & 0 & 1 \\
    \cline{2-6}
    & s.\_cov. & 1 & 1 & 0 & 0 \\
    
    \midrule
    \multirow{3}{*}{\parbox{7cm}{\textbf{Q17} Find the difference in Holdout ROC AUC between the best and the worst perf. pipelines}} & \textbf{0.231} & 8 & 6 & 2 & 0  \\
    \cline{2-6}
    & 0.138 & 1 & 0 & 0 & 1 \\
    \cline{2-6}
    & N/A & 1 & 0 & 0 & 1 \\
    
    \midrule
    & \textbf{\texttt{TOTAL}} & 170 & 108 & 33 & 29\\
    \cline{2-6}
    & \faCheck &  149 &  94 & 30 & 25 \\
    & \faClose & 21 &  14 &  3 & 4 \\
    \bottomrule
\end{tabular}%
}
\end{table}
}

\section{Discussion}

With CPC being consulted most frequently to answer our questions, we believe AutoAI visualization systems should display this component in a prominent position in the UI. 
The fact that some questions were favorably addressed using the alternative leaderboard 
suggests that other components should also be displayed.
We think next steps need to taken towards brushing and linking, to better enable the interplay of other components with CPC.
We will also explore "scaling-up" issues involving larger and more diverse collections of pipelines.

Based on 
participants' feedback, we believe we are on the right track to promote a better understanding of AutoAI systems. Some subjects, however, would like to see improvements in AutoAIViz before adding it to their job routine. We can mostly address these by placing further visual elements to increase the readability of metrics, where constraints could be indicated as bars in the background of the corresponding axes. Further, interaction to zoom to intervals of numerical axes will allow users to hide outliers distorting the view, such as shown in the axis \emph{Prediction time} in Figure \ref{fig:cpc}. Another useful feature could be a \emph{close all expanded steps} button, so less trained users can quickly reset in case they get lost.

Lastly, we found it hard to significantly increase trust among the participants. This could be due to the fact that we allowed participants to only work with the system for 15 minutes, but building trust and requires more experience. Further, subjects were exposed to a dataset they are not necessarily very familiar with. We speculate that allowing participants to work with their own dataset could have led to more trust into AutoAIViz.

\section{Conclusion}
AutoAIViz with CPC has been accepted well and significantly contributed to users' understanding of AutoAI. The system did not significantly increase trust, but we identified measures to improve towards this on the component level. Overall, CPC has proven to be of central value in opening the AutoAI blackbox, especially when augmented with other visualization techniques depending on the task, such as the leaderboard. Our study leads to future work towards enabling brushing and linking in AutoAIViz, exploring further techniques suggested for classic PC to improve CPC, and providing insights into other steps of the machine learning pipeline.


\bibliographystyle{ACM-Reference-Format}
\bibliography{sample-base}


\end{document}